\documentclass{style/hcrg-nasa}

\issuer{Harvard Computational Robotics Group}
\edition{Preprint}

\usepackage[utf8]{inputenc}
\usepackage{url}
\usepackage{array}
\usepackage{tikz}
\usetikzlibrary{arrows.meta,positioning,fit,calc}

\newcommand{\astra}{GPT-6-Astra}
\newcommand{\best}[1]{\textbf{#1}}
\title{Can a Frontier VLM Serve as a Robot Policy for Memory-Dependent Manipulation?}
\author[1*]{Jack B. Jedlicki}
\author[1]{Amish Sethi}
\author[1]{Heng Yang}
\affiliation[1]{Harvard University}
\contribution[*]{Corresponding author: \texttt{jackbjed@g.harvard.edu}}

\abstract{
We use a frozen frontier vision-language model, \astra{}, directly as the robot policy on RoboMME, a benchmark of 16
memory-dependent manipulation tasks spanning temporal, spatial, object, and procedural memory. The model keeps the whole episode in context and outputs every end-effector waypoint
and gripper command, with no RoboMME-specific training. On the 800 test episodes it succeeds in 92.1\%, above the previously
reported human score of 90.5\%, which was obtained with an oracle skill interface, and far ahead of the best
leaderboard entry (60.8\%) and of a hierarchical system that pairs the same model with a RoboMME-trained policy (79.1\%).
It matches or exceeds the human score on 13 of the 16 tasks. Its 63 failures concentrate in two tasks and are mostly
premature completions and mistimed or unintended button presses. These results blur
the usual division between long-horizon planning, memory and end-effector control, and raise the question of how much of
this stack a single general-purpose model can take over in embodied systems. Project website: \url{https://embodied-astra.github.io}. Code, prompts and all episode logs: \url{https://github.com/jackbjed/embodied-astra-robomme}.
}

\begin{document}
\maketitle

\begin{figure}[H]
\centering
\includegraphics[width=\linewidth]{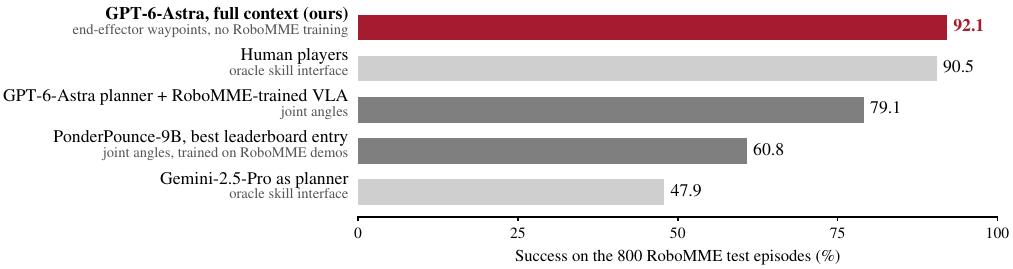}
\caption{Success on the 800 RoboMME test episodes, as reported by each source. \astra{} keeps the whole episode in context
and commands every end-effector waypoint, with no RoboMME-specific training.
Light grey bars use RoboMME's oracle skill interface, in which a privileged planner executes chosen skills. Details in Tables~\ref{tab:compare} and~\ref{tab:mc}.}
\label{fig:teaser}
\end{figure}

\section{Introduction}
Many manipulation tasks cannot be solved from the current observation alone: a robot must count how often it has done
something, track an object hidden and moved while out of view, or reproduce a motion it saw demonstrated. RoboMME
\citep{dai2026robomme} turns these requirements into 16 simulated tabletop tasks in four suites (counting, permanence,
reference and imitation). Policies on its public leaderboard remain well below its human reference
\citep{robommeleaderboard}, and in its own evaluation of VLMs as high-level planners the best model, Gemini-2.5-Pro, reaches
47.9\%, against 90.5\% for human players using the same oracle skill interface \citep{dai2026robomme}.

We ask how far a simple configuration gets on this benchmark: a frozen frontier VLM whose context window holds the whole
episode and whose outputs are the robot's actions. We use \astra{} without RoboMME-specific parameter updates and keep all
sampled observations and its previous responses in one conversation. At every step it outputs the next end-effector pose and
gripper command, which RoboMME's motion planner executes; this is RoboMME's \texttt{waypoint} action space, which the
leaderboard accepts.

Neither ingredient is new. Frozen VLMs have recently been used as direct robot policies with interaction histories
\citep{roboicl2026,unexpectedrobotpolicy2026,incontextvla2026,robodawn2026}, and on RoboMME a pretrained multimodal model
already keeps the episode in context for a learned controller \citep{choi2026ponderpounce}. A hierarchical system has
already used \astra{} to plan subtasks for a RoboMME-trained policy \citep{chen2026astrarobomme}. Our contribution is a
measurement and its explanation:
\begin{enumerate}\setlength{\itemsep}{0pt}
\item a full-context baseline, in which the frozen model directly commands Cartesian waypoints, reaches 92.1\% (737/800) in a
single run, compared with prior systems that differ in interface, training and context (Figure~\ref{fig:teaser} and Table~\ref{tab:compare});
\item an episode-level audit of all 63 failures (Figures~\ref{fig:failures} and~\ref{fig:rollouts}). In the largest groups,
the model named the demonstrated manner or explicitly stated the arrival it was waiting for, and the observable failures
are completion conditions, unintended contact and timing;
\item paired diagnostics in RoboMME's oracle-planner setting, where denser frames and a corrected prompt help \astra{} far
more than GPT-4o on the same episodes.
\end{enumerate}
We do not claim to have solved robot memory or to identify an internal memory mechanism; Section~\ref{sec:limits} lists the
controlled ablations needed for those claims.

\section{Related Work}\label{sec:related}
\paragraph{Foundation models as robot policies.}
Language and vision-language models have served as high-level planners that choose among skills
\citep{ahn2022saycan,huang2022innermonologue,qi2026composebyfocus,jedlicki2026occ4m}, write programs \citep{liang2022codeaspolicies} or compose spatial objectives
\citep{huang2023voxposer}, while vision-language-action models map observations to low-level actions after robot training
\citep{brohan2023rt2,kim2024openvla,black2024pi0,pi2025pi05}. Several recent works use a frozen frontier VLM, often \astra{}, as
the policy itself through a compact action interface. RoboICL conditions frozen \astra{} on demonstrations and a bounded
observation--action--outcome history and outputs action chunks \citep{roboicl2026}. Zhang et al.\ evaluate off-the-shelf
\astra{} on RoboDojo with end-effector target tools, keeping images from only the last two turns, and report 38.7\% success in
its memory category \citep{unexpectedrobotpolicy2026}. GPT-Policy selects Cartesian waypoint and gripper tools using task
references and online interaction history \citep{incontextvla2026}. Show-Harness and RoboDawn connect VLMs to robots through
discrete semantic or motion commands \citep{showharness2026,robodawn2026}. These works evaluate other benchmarks or real robots;
we study one benchmark designed to isolate memory, keep the full episode in context, and analyse its failures episode by
episode.

\paragraph{Memory on RoboMME.}
RoboMME's leaderboard compares memory designs for learned policies, from frame sampling and token dropping to recurrent
memory \citep{dai2026robomme}. Learned latent memories continue to be proposed \citep{mem4act2026}. PonderPounce uses a pretrained
multimodal LLM as an episode context engine for a fast controller trained on RoboMME demonstrations
\citep{choi2026ponderpounce}. Chen et al.\ use \astra{} to plan grounded subtasks that a RoboMME-trained VLA executes, with a
learned completion monitor \citep{chen2026astrarobomme}. Table~\ref{tab:compare} sets these systems side by side. They differ
in more than one factor, so the score differences cannot be attributed to the model, the context policy or the interface alone.

\begin{table}[!tbp]
\centering
\footnotesize
\setlength{\tabcolsep}{3pt}
\begin{tabular}{>{\raggedright}p{2.1cm}>{\raggedright}p{2.8cm}>{\raggedright}p{2.1cm}>{\raggedright}p{1.8cm}>{\raggedright}p{1.9cm}cc}
\toprule
System & Decision model & Action interface & RoboMME-trained parameters & Status & Runs & Success \tabularnewline
\midrule
Human players \citep{dai2026robomme} & 18 participants & skill selection, oracle planner & -- & reference & 1 & 90.5 \tabularnewline
\midrule
PonderPounce-9B \citep{choi2026ponderpounce} & Qwen3.5-9B context engine + $\pi_{0.5}$-initialized controller & joint angles & yes (1,587 demos) & leaderboard & 3 & 60.8 \tabularnewline
\quad 9$\times$ data & same & joint angles & yes (14,400 episodes) & paper only & 3 & 75.5 \tabularnewline
Three-tier \citep{chen2026astrarobomme} & \astra{} planner + RoboMME VLA + learned monitor & joint angles (VLA) & yes (VLA, monitor) & project report & 1 & 79.1 \tabularnewline
\midrule
Ours & \astra{}, full episode context & Cartesian waypoints, RoboMME motion planner & none & leaderboard & 1 & 92.1 \tabularnewline
\bottomrule
\end{tabular}
\caption{Results on the 800 RoboMME test episodes (\%), as reported by each source; leaderboard snapshot of 2026-09-29,
accessed 2026-10-10 \citep{robommeleaderboard}. Only the human reference uses the oracle skill interface; all other rows
issue low-level or waypoint actions. RoboMME's leaderboard also lists a baseline that executes ground-truth subgoals (84.1\%).}
\label{tab:compare}
\end{table}

\section{Setup}\label{sec:method}

\paragraph{Episodes and observations.}
Each RoboMME episode provides a language goal $g$ and, for 9 of the 16 tasks, a video $V$ that specifies the task or shows a
scene that later changes (a demonstration, or objects that are later hidden). At decision step $i$ the simulator returns the
frames recorded since the previous decision, $F_i=(f_{i,0},\dots,f_{i,n_i-1})$, from a fixed front camera (256$\times$256),
together with a wrist-camera image and the robot's proprioception. The simulator records one frame per control step, at
20~steps per second of simulated time.

\paragraph{Context.}
We subsample each clip $F=(f_0,\dots,f_{n-1})$ with stride $k$,
\begin{equation}
S_k(F)=\{f_j : j \equiv 0 \!\!\pmod k\}\cup\{f_{n-1}\},
\end{equation}
always including the latest frame, and condition the model on all sampled observations and its previous responses,
\begin{equation}
a_i \sim \pi\big(\,\cdot \mid g,\; S_k(V),\; S_k(F_1),\, y_1,\; \dots,\; S_k(F_i)\big),
\end{equation}
where $y_j$ is the model's answer at step $j$, which includes a free-text description of what happened as well as the action.
The policy $\pi$ is \astra{} queried through the Responses API, with all turns of an episode chained in one conversation.
The harness uses the API's default sampling settings, which every response records as medium reasoning effort,
temperature 1, top-$p$ 0.98 and truncation disabled.
Apart from the subsampling $S_k$, nothing is summarized or dropped. With $k=3$, the model sees one front-camera frame every
0.15~s of simulated time; with $k=30$, one every 1.5~s. RoboMME's VLM pipeline encodes the frames as 30~fps video and
samples it at 1~frame/s for Gemini, so its effective spacing is also 30 frames, or 1.5~s of simulated time.

\paragraph{Waypoint interface.}
Each action is a waypoint $a_i=(x,r,\gamma)$: a tool-center-point (TCP) position $x\in\mathbb{R}^3$, an orientation $r$ (roll,
pitch, yaw) and a gripper command $\gamma\in\{-1,0,1\}$ ($-1$ close, $0$ keep, $1$ open), all in the world frame. RoboMME's
motion planner moves the arm to the commanded pose, with no snapping to objects and no privileged state. This is control at
the level of end-effector waypoints, not joint or actuator commands. Besides the images, the model receives the TCP pose,
finger positions, elapsed control steps and the camera intrinsics and extrinsics, which RoboMME exposes as optional
observations; it never receives depth, privileged object state or the benchmark's oracle subgoal annotations. Its system prompt, ours since
RoboMME provides none for this setting, is shared by all tasks apart from one sentence stating the robot's hardware in
PatternLock and RouteStick (it holds a stick, so the gripper command is ignored). The prompt describes the action format, the
world frame, general properties of the robot interface (motion timing, gripper behaviour) and one convention: left and right
in task instructions are the robot's, and the front camera, which faces the robot, shows them mirrored. RoboMME's
instructions do not say whose left is meant, while its own oracle-planner prompts state such conventions for individual
tasks; we state it once for all tasks. We use stride $k=3$ for the front camera and add the current wrist-camera frame at
each step.

\begin{figure}[!tbp]
\centering
\resizebox{\linewidth}{!}{%
\begin{tikzpicture}[
  font=\small, >={Stealth[length=2.2mm]},
  box/.style={draw, rounded corners=2pt, align=center, inner sep=4pt, minimum height=11mm},
  lab/.style={font=\footnotesize, text=black!75, align=center}]
\node[box, minimum width=27mm] (env) {RoboMME task\\{\footnotesize ManiSkill simulator}};
\node[box, fill=black!6, minimum width=34mm, minimum height=15mm, right=32mm of env] (llm)
  {\textbf{\astra{}}\\{\footnotesize context = episode history}};
\coordinate (split) at ($(llm.east)+(9mm,0)$);
\node[box, minimum width=37mm, anchor=west] (wp) at ($(split)+(30mm,10mm)$)
  {Motion planner\\{\footnotesize move to pose, apply gripper}};
\node[box, minimum width=37mm, anchor=west] (mc) at ($(split)+(30mm,-10mm)$)
  {Oracle skill planner\\{\footnotesize skill $c$ on object nearest $p$}};
\draw[->] (env) -- node[lab, above] {goal, task video} node[lab, below] {frames $S_k(F_i)$, state} (llm);
\draw (llm.east) -- (split);
\draw[->] (split) |- node[lab, above, pos=0.75] {waypoint $(x,r,\gamma)$} (wp.west);
\draw[->] (split) |- node[lab, below, pos=0.75] {oracle planner $(c,p)$} (mc.west);
\coordinate (ret) at ($(mc.east)+(5mm,0)$);
\draw (wp.east) -| (ret);
\draw (mc.east) -- (ret);
\draw[->] (ret) -- ++(0,-12mm) -| node[lab, below, pos=0.25] {new frames after each executed action} (env.south);
\end{tikzpicture}}
\caption{The model receives the goal, the task video and subsampled frames of everything that happens, and keeps its own
previous answers in context. In the waypoint setting (leaderboard) it commands every end-effector pose; in RoboMME's
oracle-planner setting (diagnostic only) it selects a skill that an oracle executes.}
\label{fig:system}
\end{figure}
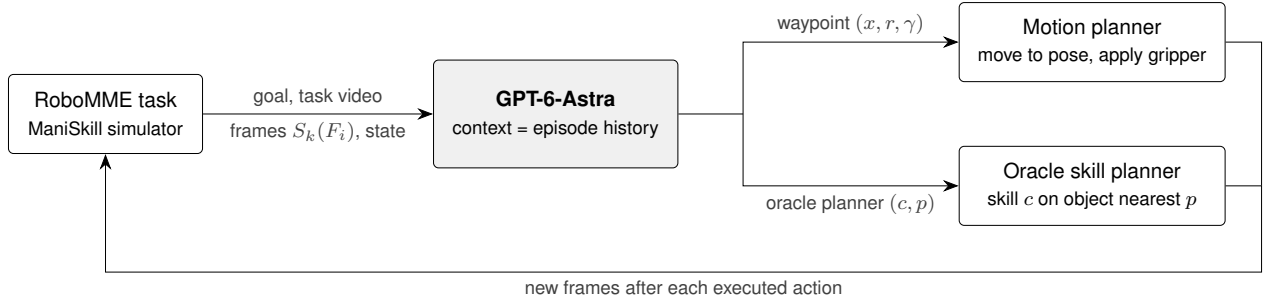

\paragraph{Oracle-planner interface.}
For comparison with RoboMME's VLM evaluation, the model can instead output $a_i=(c,p)$: a skill $c$ from the task's fixed
option list and an image point $p\in[0,1000]^2$. RoboMME's oracle planner then executes $c$ on the object nearest $p$ using
privileged simulator state. This is the setting of RoboMME's Table~11 and of its human players. The leaderboard does not accept
it. Here we use RoboMME's own task prompts and query templates, the front camera only, and $k=30$
(Figure~\ref{fig:system}).

\paragraph{Protocol.}
We follow RoboMME's evaluation script: the 50 test episodes of each task with their fixed seeds, a limit of 1300 control
steps, and success as judged by the environment, with one run per episode. The main configurations cover all 800 episodes; the oracle-setting corrections of
Section~\ref{sec:oracle} rerun only the tasks they affect and reuse the outcomes of the others. Simulator errors and unusable answers count as
failures, as do episodes in which the model declares the task complete before the environment judges success, three
consecutive actions that execute no steps, and a cap of 150 model calls (never reached; the maximum was 31). No episode is
excluded or retried. We use RoboMME at commit \texttt{856bc3a}. We report 95\% Wilson intervals and compare configurations on the same episodes with exact McNemar tests.
These intervals describe sampling over episodes only. They do not account for adaptive development on the test split
(Section~\ref{sec:limits}), for API drift, or for generalization to new tasks.

\section{Waypoint-Policy Results}\label{sec:results}

\begin{table}[!tbp]
\centering
\small
\setlength{\tabcolsep}{5pt}
\begin{tabular}{llccc}
\toprule
Suite & Task & Human$^\dagger$ & PP-9B$^\ddagger$ & \astra{} \\
\midrule
\multirow{4}{*}{Counting}
 & BinFill          & 96  & 50.7 & \best{98} \\
 & PickXtimes       & 100 & 93.3 & \best{100} \\
 & SwingXtimes      & 80  & 72.0 & \best{98} \\
 & StopCube         & 78  & 82.7 & 54 \\
\midrule
\multirow{4}{*}{Permanence}
 & VideoUnmask      & 90 & 94.7 & \best{94} \\
 & VideoUnmaskSwap  & 92 & 25.3 & \best{96} \\
 & ButtonUnmask     & 92 & 97.3 & \best{100} \\
 & ButtonUnmaskSwap & 90 & 34.0 & \best{98} \\
\midrule
\multirow{4}{*}{Reference}
 & PickHighlight    & 92 & 85.3 & \best{98} \\
 & VideoRepick      & 92 & 40.7 & \best{100} \\
 & VideoPlaceButton & 98 & 83.3 & \best{100} \\
 & VideoPlaceOrder  & 90 & 79.3 & \best{98} \\
\midrule
\multirow{4}{*}{Imitation}
 & MoveCube         & 90 & 81.3 & 54 \\
 & InsertPeg        & 98 & 6.0  & 92 \\
 & PatternLock      & 84 & 12.7 & \best{100} \\
 & RouteStick       & 86 & 34.7 & \best{94} \\
\midrule
 & \textbf{Average} & 90.5 & 60.8 & 92.1 \\
\bottomrule
\end{tabular}
\caption{Success rate (\%) on the 50 test episodes of each task. $^\dagger$Previously reported human-reference score
\citep{dai2026robomme}, obtained with the oracle skill interface rather than waypoints. $^\ddagger$PonderPounce-9B leaderboard
entry, joint-angle control, mean of 3 runs \citep{choi2026ponderpounce}. \astra{} is a single run. Bold marks scores at or above the
human reference.}
\label{tab:e2e}
\end{table}

\begin{figure}[!tbp]
\centering
\includegraphics[width=\linewidth]{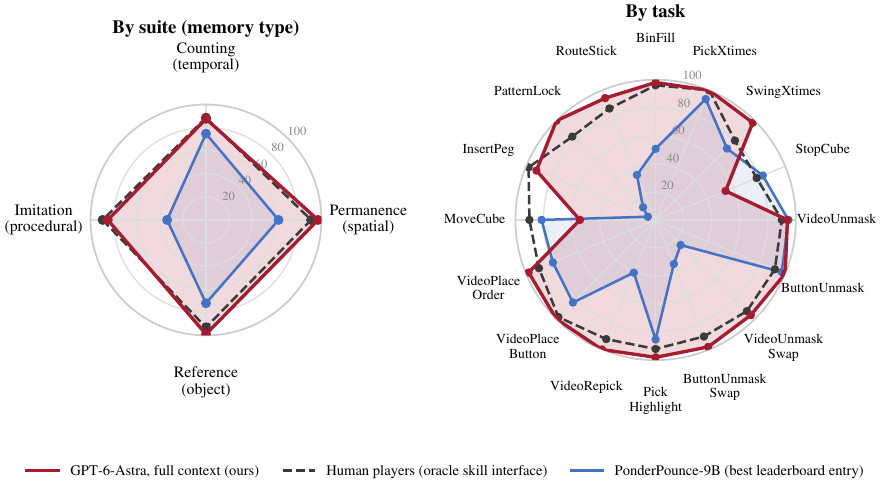}
\caption{Success rate (\%) on the RoboMME test split by suite, with the memory type each suite targets (left; mean of its
four tasks), and by task (right). \astra{} is a single run; the human score was obtained with the oracle skill interface;
PonderPounce-9B is the best leaderboard entry (joint-angle control, mean of 3 runs). Values as in Table~\ref{tab:e2e}.}
\label{fig:radar}
\end{figure}

\paragraph{Overall.}
Commanding the end effector itself, \astra{} succeeds in 737 of 800 episodes, 92.1\% [90.1, 93.8] (Table~\ref{tab:e2e}). This
is 31.3 points above the best leaderboard entry (60.8\%), a learned policy trained on RoboMME demonstrations. The same
entry's authors report 75.5\% with nine times more training data \citep{choi2026ponderpounce}, and the three-tier system
that uses \astra{} as a planner reports 79.1\% \citep{chen2026astrarobomme}, 13.0 points below our run. Because these systems
also differ in execution stack, training and number of runs, these gaps measure systems rather than any single design
choice. The aggregate is numerically above the previously reported human reference of 90.5\%, which lies inside our
interval and was obtained with a different interface. Per task, the score is at or above the human reference on 13 of 16
tasks: three counting tasks, every permanence and reference task, and two imitation tasks, PatternLock and RouteStick. On
these two and on InsertPeg, where it reaches 92\% against 98\% for humans, the best leaderboard entry reaches 6.0--34.7\%. Two tasks account for 46 of the 63 failures: MoveCube and StopCube (54\% each). Figure~\ref{fig:radar} shows the same results by suite and by task.

\paragraph{Resources.}
The model made 7{,}563 calls, 9.5 per episode on average (median 8, 95th percentile 20, maximum 31). Each call took 18.3~s on
average as measured by the client. The server-reported generation time had a median of 14~s, a 90th percentile of 33~s, a
95th percentile of 45~s and a maximum of 77~s. An episode used a median of 148 images (maximum 518). Its final request carried
a median of 16.5k input tokens (maximum 49.9k). From the API-reported token counts, at \$10, \$1, \$12.50 and \$50 per million
fresh-input, cached-input, cache-write and output tokens, the run cost \$309 (\$0.39 per episode); nearly all input tokens
were cache reads or writes. The simulator waits during each call, so this is a capability baseline, not a real-time
controller.

\section{Failure Analysis}\label{sec:failures}

\begin{figure}[!tbp]
\centering
\includegraphics[width=\linewidth]{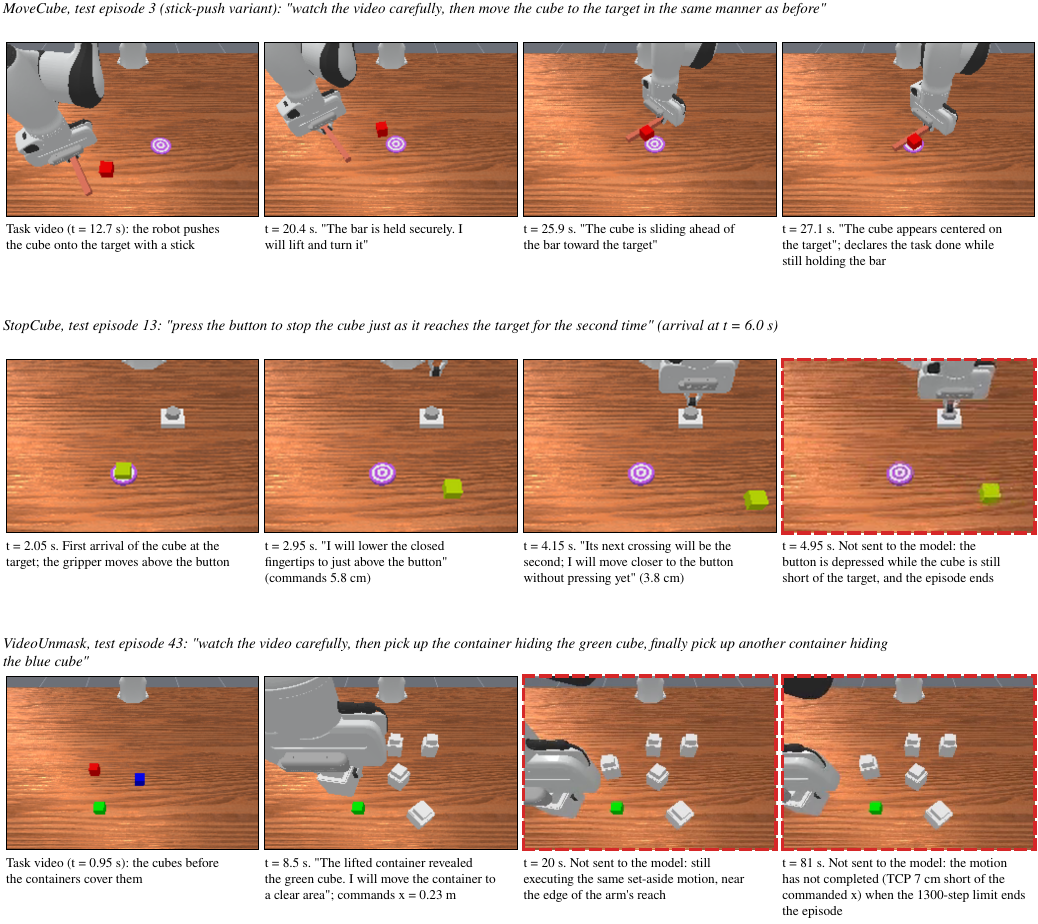}
\caption{Recorded front-camera frames of failed rollouts from the three most frequent groups; $t$ is simulated time and
quotes are the model's own descriptions. Frames with a plain border were sent to the model as observations; frames with a
dashed red border are execution or terminal frames from the episode recording that the model never received, because the
episode ended before its next request. Top: MoveCube (stick-push variant), premature completion while
still holding the stick. Middle: StopCube, unintended button contact while positioning. Bottom: VideoUnmask, a set-aside
waypoint beyond the arm's reach that the executor runs until the step limit.}
\label{fig:rollouts}
\end{figure}

\paragraph{Labelling.}
We label each of the 63 failed episodes with one primary class from its records: the logged terminal reason, the executed
waypoints and their outcomes, the model's own descriptions, and task parameters read back from the environment after reset
(the cube's schedule in StopCube, the MoveCube variant). Classes describe observable behaviour, not internal causes; no
class asserts a ``memory failure''. Every failure received exactly one class under the rules listed in
Appendix~\ref{app:audit}. Figure~\ref{fig:failures} shows the result, and Figure~\ref{fig:rollouts} shows one recorded rollout
for each of the three most frequent tasks.

\begin{figure}[!tbp]
\centering
\includegraphics[width=\linewidth]{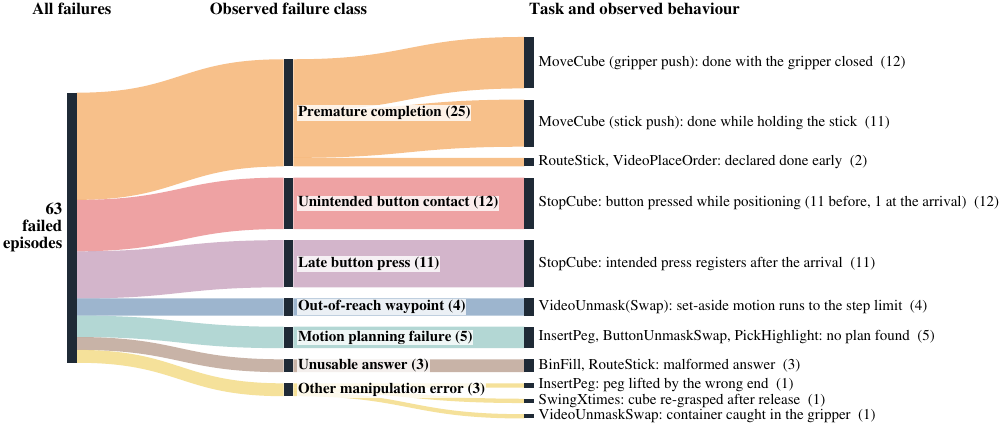}
\caption{All 63 failures of the run (737 of 800 episodes succeeded), by observed class and by the task and behaviour in which
they occur. Widths are proportional to episode counts. Classes are assigned from episode records
(Appendix~\ref{app:audit}); they describe behaviour, not internal causes.}
\label{fig:failures}
\end{figure}

\paragraph{Premature completion (25).}
In 25 episodes the model declared the task complete before the environment judged success. Twenty-three are in MoveCube,
which has three variants: grasping and placing the cube, pushing it with the closed gripper, or pushing it with a stick. The
model solved all 21 grasp-and-place episodes but only 5 of 17 gripper-push and 1 of 12 stick-push episodes. RoboMME's success
check for the pushing variants requires the gripper to be open at the end. In all 23 failures, the model's first description
names the demonstrated manner (pushing with the bar, or sliding the cube along the table without lifting it), and its last
gripper command before declaring completion was to close: in 11 stick-push episodes it still held the stick
(Figure~\ref{fig:rollouts}, top), and in 12 gripper-push episodes the gripper was still closed. The remaining two are one
RouteStick and one VideoPlaceOrder episode.

\paragraph{Unintended button contact (12) and late press (11).}
StopCube asks the robot to press a button so that a cube oscillating across a target stops on it at a given arrival. The
model's last action separates two behaviours among the 23 failures. In 12, it was described as positioning (``move closer to
the button without pressing yet''), with a commanded TCP height of 3.2--4.0~cm, yet the fingertips depressed the button
(Figure~\ref{fig:rollouts}, middle); in 11 of these the press registered before the required arrival, read from the
environment, and in one exactly at it. In the other 11, the model intended to press, with commanded heights of
0.8--1.8~cm, and the press registered after the arrival. In all 23 failures, the model's last description names an arrival
(``its next crossing will be the second''). We did not check these statements against the cube's state, so they show what
the model intended to do, not that its count was right. The dominant StopCube failures are therefore, as observed behaviour, unintended contact and mistimed
presses; because the stated arrivals were not checked, these observations do not rule out counting or perception errors.

\paragraph{Out-of-reach waypoint (4) and motion planning failures (5).}
In four VideoUnmask(Swap) episodes the model lifted the correct first container and commanded a set-aside pose far forward
($x=0.22$--$0.23$~m, $|y|=0.23$--$0.25$~m). RoboMME's executor did not report the pose as unreachable: it ran for 1{,}096--1{,}453
control steps, the arm stopping 6--7~cm short of the commanded $x$ (Figure~\ref{fig:rollouts}, bottom), and the episodes
reached the 1300-step limit. Across the run, 10 of the 11 executed waypoints with $x\ge0.18$~m took 581--1{,}453 steps each; five of
those ten episodes still succeeded. In five other episodes (three InsertPeg, one ButtonUnmaskSwap, one PickHighlight) the planner
reported that no motion plan to the commanded pose was found.

\paragraph{Unusable answers (3) and other manipulation errors (3).}
Three answers could not be parsed: two contained a number partly written in words (\texttt{0.\ seventeen},
\texttt{-0.\ eighteen}); the third contained a digit in another script and then text imitating a harness message. The other
three failures are single cases: in InsertPeg the model lifted the demonstrated peg by its other end, which the environment
treats as a failure; in SwingXtimes, with left and right correct, the cube slid off the target after release and the model picked it up
again, which the environment judged a failure; in VideoUnmaskSwap the first container stayed caught between the fingers
after a long set-aside motion, and the episode failed 16 actions later, after repeated attempts to release it and grasp the
second container.

\section{Diagnostics in RoboMME's Oracle-Planner Setting}\label{sec:oracle}

\paragraph{Calibration.}
To compare with RoboMME's VLM evaluation, we run GPT-4o (\texttt{gpt-4o-2024-08-06}) through our oracle-planner harness with
RoboMME's prompts and seeds. It reaches 30.8\% [27.7, 34.0], against 30.6\% reported by RoboMME, so the harness approximately
recovers the reported aggregate. The protocols still differ. RoboMME's released code sampled GPT-4o's frames more densely
than we do (every 8th frame, at most 20 per message, against one in 30), and we count invalid actions as failures. Per-task
differences exceed sampling noise on StopCube and SwingXtimes, and the agreement of the aggregates does not establish
identical evaluation semantics.

\paragraph{\astra{} as a planner.}
With execution delegated to the oracle, \astra{} succeeds in 77.8\% [74.7, 80.5] of episodes, against 47.9\% for
Gemini-2.5-Pro, 30.6\% for GPT-4o and 27.9\% for Qwen3-VL-32B in RoboMME's evaluation (Table~\ref{tab:mc}; per-task results in
Appendix~\ref{app:mc}). This number is not directly comparable to the waypoint result. The two settings differ in executor,
in frame density ($k=30$ vs.\ $k=3$) and in access to the wrist camera, so 77.8\% versus 92.1\% does not isolate the
difficulty of control. Its weakest tasks fall into two groups. On tasks that hinge on brief events (container swaps, a cube
re-picked from a shuffled set, demonstrated stick trajectories), one frame per 1.5~s misses information. On PickHighlight
and StopCube, the failures follow from the oracle interface. RoboMME's prompt says the gripper reopens after a button press,
which the oracle's press skill never does, so the model repeats the press. On StopCube, the oracle's ``remain static'' skill
waits only until a fixed checkpoint just before the press time and then stops advancing time.

\paragraph{Interface corrections.}
We evaluate three corrections as labelled variants:
\textbf{D}, denser frames ($k=3$ instead of $30$, as in the waypoint setting), on the five tasks above;
\textbf{G}, a note in the PickHighlight prompt that the press-button skill ends with the gripper closed;
\textbf{F+T}, for StopCube and together with D, a message when an action executes no steps and a note that pressing takes
about 1.5~s and that the button must be pressed once waiting has no effect. F+T is an \emph{oracle-interface diagnostic}, not
a neutral fix. The oracle's waiting skill stops a fixed 1.5~s before the cube must be stopped, which is exactly when the
button must be pressed, and the note's 1.5~s is that interval, so F+T passes timing information through the executor.
RoboMME's released human-evaluation interface uses the same waiting skill. D was evaluated directly on the test episodes. G
and F+T were first checked on RoboMME's separate validation episodes (PickHighlight 2/10$\to$10/10, StopCube
1/10$\to$10/10) and then run once on the test set. With D and G the model reaches 92.3\% [90.2, 93.9]; adding F+T gives 94.6\%
[92.8, 96.0], a gain that should not be read as evidence of visual timing.

\begin{table}[!tbp]
\centering
\small
\begin{tabular}{lccccc}
\toprule
Setting & \astra{} & GPT-4o & Gemini-2.5-Pro & Qwen3-VL-32B & Human \\
\midrule
RoboMME's evaluation \citep{dai2026robomme} & -- & 30.6 & 47.9 & 27.9 & 90.5 \\
Our harness                                 & 77.8 & 30.8 & -- & -- & -- \\
\quad + D, G                                & 92.3 & 31.9 & -- & -- & -- \\
\quad + D, G, F+T (diagnostic)              & 94.6 & 32.4 & -- & -- & -- \\
\bottomrule
\end{tabular}
\caption{Oracle-planner setting: average success rate (\%) over the 800 test episodes. Corrections apply to the tasks of
Table~\ref{tab:paired} only; all other tasks are unchanged.}
\label{tab:mc}
\end{table}

\begin{table}[!tbp]
\centering
\small
\setlength{\tabcolsep}{4.5pt}
\begin{tabular}{llcccccc}
\toprule
 & & \multicolumn{3}{c}{\astra{}} & \multicolumn{3}{c}{GPT-4o} \\
\cmidrule(lr){3-5}\cmidrule(lr){6-8}
Variant & Tasks & base & variant & gained / lost & base & variant & gained / lost \\
\midrule
D   & 5 video-memory tasks & 159 & 239 & 83 / 3 & 32 & 37 & 19 / 14 \\
G   & PickHighlight        & 14  & 50  & 36 / 0 & 20 & 24 & 7 / 3 \\
\midrule
\multicolumn{2}{l}{D and G (300 episodes)} & 173 & 289 & 119 / 3 & 52 & 61 & 26 / 17 \\
\multicolumn{2}{l}{McNemar exact $p$} & \multicolumn{3}{c}{$1\times10^{-31}$} & \multicolumn{3}{c}{$0.22$} \\
\midrule
F+T & StopCube (with D, diagnostic) & 24 & 43 & 23 / 4 & 21 & 25 & 12 / 8 \\
\bottomrule
\end{tabular}
\caption{Paired control in the oracle-planner setting. Successful episodes (of 50 per task) without and with each
correction, on the same test episodes. ``Gained / lost'' counts episodes that switch outcome.}
\label{tab:paired}
\end{table}

\begin{figure}[!tbp]
\centering
\includegraphics[width=0.92\linewidth]{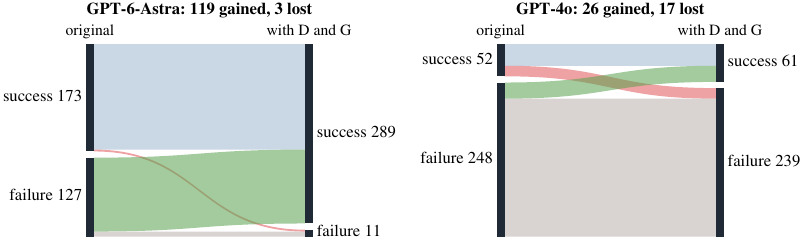}
\caption{Paired outcomes of the 300 test episodes affected by D and G, for \astra{} and GPT-4o, drawn to the same scale. Unlike
two aggregate bars, the flows show both recovered failures (green) and new failures (red). F+T is not included.}
\label{fig:paired}
\end{figure}

\paragraph{Paired control.}
A correction derived from one model's failures might help every model. We therefore apply identical corrections to GPT-4o on
the same episodes (Table~\ref{tab:paired}, Figure~\ref{fig:paired}). With D and G, \astra{} gains 119 episodes and loses 3
($p=1\times10^{-31}$). GPT-4o gains 26 and loses 17 ($p=0.22$), a change that is not statistically significant in this
sample; non-significance does not show that the corrections have no effect on GPT-4o. To test the difference between the
two models directly, we estimate the interaction
$\Delta=(S_{\text{A,corr}}-S_{\text{A,base}})-(S_{\text{4o,corr}}-S_{\text{4o,base}})$ over the 300 affected episodes:
$\hat\Delta=35.7$ points, with a 95\% percentile interval of [29.0, 42.3] from a task-stratified bootstrap that resamples
episodes jointly across both models and both conditions (20{,}000 resamples, fixed seed). The per-episode outcomes of all
oracle-planner runs and the script that recomputes Tables~\ref{tab:mc}, \ref{tab:paired} and~\ref{tab:mcfull} and this
interval are released with the code. The corrections help the model that can use the additional frames and the
corrected description far more than the weaker model, which argues against their simply supplying answers. F+T is the
exception: it carries the oracle's timing, and we report it only as a diagnostic.

\section{Limitations and Next Experiments}\label{sec:limits}
\paragraph{What the evidence does not show.}
The model sees past images, its own earlier descriptions and the record of its actions. Our experiments do not separate
these channels, so they do not show which carries the information that history-dependent tasks need, nor do they identify a
memory mechanism. The controlled test is to hold the task specification (goal and task video) fixed and remove online history
selectively: latest observation only, visual history without earlier descriptions, textual history without older images, and
a recent window. Each condition must rebuild its requests from its permitted records, so that removed history cannot persist
through a chained conversation. Counterfactual episodes with identical current observations but different relevant histories,
and tasks that vary delay and interference, would further separate memory demands from current-state shortcuts.

\paragraph{Confounds.}
The waypoint and oracle-planner settings differ in executor, frame density and camera access; a factorial comparison of
\{waypoint, oracle\} $\times$ \{dense, sparse\} with matched observations would isolate the executor. No other model was evaluated
under the waypoint harness, so the interface notes and the left/right convention in its prompt are not paired with another
model in that setting.

\paragraph{Development, reproducibility and scope.}
Configuration choices were made during development that included evaluations on the test split: frame density, the interface
notes in the system prompt, the oracle-setting corrections, and a pilot of an earlier harness version on two test episodes.
In particular, the left/right convention was added to the shared prompt after an earlier full run of the same harness
without it, in which the model failed all 50 SwingXtimes episodes by reading left and right from the mirrored front image;
the note was then run on SwingXtimes alone, on 10 validation episodes (9/10) and on its 50 test episodes (50/50), before
the full run reported here. The logs of the earlier run and of these checks are at
\url{https://github.com/jackbjed/astra-robomme}. On the other 15 tasks the two runs differ in 37 of 750 episodes, in both
directions (13 gained, 24 lost; exact McNemar $p=0.10$); since the runs differ in the prompt as well as in sampling, this
mixes the effect of the note with sampling variability.

We made no RoboMME-specific parameter updates, but \astra{}'s pretraining data are unknown. The model is served through an
API without dated snapshots (we used the identifier \texttt{gpt-6-astra} on 2026-10-11), so exact reproduction depends on its
availability, and costs are computed from token counts at assumed prices, not billing records. RoboMME's tasks are simulated
tabletop tasks with a fixed camera; longer horizons, mobile manipulation and physical robots remain open.

\section{Conclusion}
A frozen frontier VLM that keeps the whole episode in context and commands Cartesian waypoints directly is a strong baseline
for RoboMME. In a single run it succeeds in 92.1\% of test episodes, above the previously reported human score of 90.5\%
obtained with an oracle skill interface, and above prior systems that train on RoboMME data. Its failures concentrate in
MoveCube and StopCube. In MoveCube, the model named the demonstrated way of moving the cube but declared completion with the
gripper closed, which the success check of the pushing variants does not accept; in StopCube, it pressed the button too late
or touched it while positioning, although its descriptions named the intended arrival, which we did not verify against the
cube's state. In the oracle-planner setting, denser frames and a corrected prompt help \astra{} far more than GPT-4o. These
results point to the need for harder memory benchmarks that push vision-language models to their limits as embodied agents,
for example with mobile manipulation, changing viewpoints and more demanding physical interaction. They also raise the
question of where to draw the boundaries between high-level planning, memory and low-level control. Code, prompts, configuration and per-episode logs for the waypoint evaluation, the failure-audit script, and the per-episode outcomes of the oracle-planner runs with the script that recomputes their statistics are available at
\url{https://github.com/jackbjed/embodied-astra-robomme}; the images sent to the model are available on request.

\bibliographystyle{plainnat}
\bibliography{refs}

\newpage
\appendix
\section{Failure Audit Rules}\label{app:audit}
Each failed episode of the run receives one primary class, assigned in the following order.
\begin{itemize}\setlength{\itemsep}{0pt}
\item \emph{Unusable answer}: terminal reason \texttt{unparseable} (BinFill ep.~30, RouteStick ep.~20 and~47).
\item \emph{Motion planning failure}: terminal reason \texttt{env\_error}; all five report that motion planning was exhausted
(ButtonUnmaskSwap ep.~11, InsertPeg ep.~2, 11 and~16, PickHighlight ep.~39).
\item \emph{Premature completion}: the model returned no waypoint and declared the task complete before the environment judged
success (23 MoveCube episodes, RouteStick ep.~24, VideoPlaceOrder ep.~46). MoveCube cases are split by the variant read from the
environment: 11 stick-push episodes (3, 4, 5, 8, 11, 15, 25, 31, 32, 33, 47) and 12 gripper-push episodes (7, 12, 14, 20, 22,
23, 26, 29, 37, 38, 41, 49); in all 23 the last gripper command was to close.
\item \emph{Out-of-reach waypoint}: the episode reached the step limit after a single waypoint that ran for more than 1{,}000
control steps and ended more than 5~cm short of its commanded position (VideoUnmask ep.~11, 43 and~47, VideoUnmaskSwap ep.~39).
\item \emph{Unintended button contact} and \emph{late press}: StopCube episodes judged failed by the environment, which ends an
episode when the button registers a press while the cube is not on the target. The class follows the model's last action:
described as positioning, with a commanded TCP height of 3.2--4.0~cm, is \emph{unintended contact} (ep.~0, 2, 9, 11, 12,
13, 14, 25, 27, 32, 35, 40); described as a press, with 0.8--1.8~cm, is a \emph{late press} (ep.~1, 5, 7, 18, 21, 24, 30, 36,
42, 44, 45). Against the arrival step read from the environment, every late press registered after it and every unintended
contact before it, except ep.~11, which registered exactly at it.
\item \emph{Other manipulation error}: the remaining environment-judged failures, each described in
Section~\ref{sec:failures} (InsertPeg ep.~30, SwingXtimes ep.~44, VideoUnmaskSwap ep.~0).
\end{itemize}
The classes are exhaustive and exclusive for this run (3+5+25+4+12+11+3 = 63). Counts are reproduced by
\texttt{analysis/audit.py} in the code repository from the released episode records and the task parameters read back from
the environment.

\section{Per-Task Oracle-Planner Results}\label{app:mc}

\begin{table}[H]
\centering
\small
\setlength{\tabcolsep}{5pt}
\begin{tabular}{llcccccc}
\toprule
Suite & Task & Human & Gemini$^\dagger$ & GPT-4o & \astra{} & +corr. & corr. \\
\midrule
\multirow{4}{*}{Counting}
 & BinFill          & 96  & 78 & 58 & \best{98}  & 98  & -- \\
 & PickXtimes       & 100 & 80 & 56 & 96         & 96  & -- \\
 & SwingXtimes      & 80  & 58 & 18 & \best{100} & 100 & -- \\
 & StopCube         & 78  & 32 & 42 & 48         & 86  & F+T$^*$ \\
\midrule
\multirow{4}{*}{Permanence}
 & VideoUnmask      & 90 & 86 & 34 & \best{100} & 100 & -- \\
 & VideoUnmaskSwap  & 92 & 18 & 20 & 78         & 100 & D \\
 & ButtonUnmask     & 92 & 82 & 58 & \best{100} & 100 & -- \\
 & ButtonUnmaskSwap & 90 & 12 & 32 & 64         & 100 & D \\
\midrule
\multirow{4}{*}{Reference}
 & PickHighlight    & 92 & 78 & 40 & 28         & 100 & G \\
 & VideoRepick      & 92 & 28 & 4  & 36         & 78  & D \\
 & VideoPlaceButton & 98 & 50 & 16 & 86         & 86  & -- \\
 & VideoPlaceOrder  & 90 & 60 & 18 & \best{90}  & 90  & -- \\
\midrule
\multirow{4}{*}{Imitation}
 & MoveCube         & 90 & 66 & 76 & 86         & 86  & -- \\
 & InsertPeg        & 98 & 32 & 12 & 94         & 94  & -- \\
 & PatternLock      & 84 & 4  & 6  & 70         & 100 & D \\
 & RouteStick       & 86 & 2  & 2  & 70         & 100 & D \\
\midrule
 & \textbf{Average} & 90.5 & 47.9 & 30.8 & 77.8 & 94.6 & \\
\bottomrule
\end{tabular}
\caption{Oracle-planner success rate (\%) on the 50 test episodes of each task. Human and Gemini-2.5-Pro ($^\dagger$) are as
reported \citep{dai2026robomme}; GPT-4o is our run in the same harness (RoboMME reports 30.6\%). ``+corr.'' applies the
correction in the last column to that task only. $^*$F+T is an oracle-interface diagnostic; without it the average is
92.3\%. Bold marks \astra{} results at or above the human score.}
\label{tab:mcfull}
\end{table}

\end{document}